\documentclass[runningheads]{llncs}  

\usepackage{graphics} 
\usepackage{epsfig} 
\usepackage{amsmath} 
\usepackage{amssymb}  
\usepackage{comment}
\usepackage{bm}
\usepackage{booktabs,lipsum,calc}
\usepackage{color}
\usepackage{IEEEtrantools}

\begin{document}
	
\title{
A Novel Approach to Forecasting Financial Volatility with Gaussian Process Envelopes 
}
\titlerunning{Forecasting Financial Volatility with Gaussian Process Envelopes}
\author{Syed Ali Asad Rizvi \and Stephen J. Roberts \and \\ Michael A. Osborne \and Favour Nyikosa}
\authorrunning{S.A.A Rizvi et al.}
\institute{Machine Learning Research Group, \\Oxford-Man Institute of Quantitative Finance, University of Oxford \\
	\email{\{arizvi,sjrob,mosb,favour\}@robots.ox.ac.uk}}

\maketitle

\begin{abstract}
	
	In this paper we use Gaussian Process (GP) regression to propose a novel approach for predicting volatility of financial returns by forecasting the envelopes of the time series. We provide a direct comparison of their performance to traditional approaches such as GARCH. We compare the forecasting power of three approaches: GP regression on the absolute and squared returns; regression on the envelope of the returns and the absolute returns; and regression on the envelope of the negative and positive returns separately. We use a maximum a posteriori estimate with a Gaussian prior to determine our hyperparameters. We also test the effect of hyperparameter updating at each forecasting step. We use our approaches to forecast out-of-sample volatility of four currency pairs over a 2 year period, at half-hourly intervals. From three kernels, we select the kernel giving the best performance for our data. We use two published accuracy measures and four statistical loss functions to evaluate the forecasting ability of GARCH vs GPs. In mean squared error the GP's perform 20\% better than a random walk model, and 50\% better than GARCH for the same data. 
	
\end{abstract}


\section{Introduction}

In financial time series, volatility is a measure of the uncertainty present in the market at a given point in time. Financial time series are understood to be heteroskedastic in nature, i.e.\ the volatility varies with time. The volatility also exhibits a phenomenon known as bunching or volatility clustering, where periods of high volatility are often followed by periods of high volatility and periods of low volatility usually follow periods of low volatility. Returns of a financial time series are price movements and are understood to be a measure of the difference between two consecutive values of the time series. Returns are more precisely defined in the next section. Negative and positive returns do not affect volatility in the same way. It is often seen that large negative returns seem to raise volatility more than positive returns which have the same absolute value i.e the market reacts with more volatile and risky behaviour to the downward trend of the price of an asset, and with more cautious optimism to the upward trend of an asset price. 

Volatility forecasting is important for various types of economic agents, ranging from option traders to price options, to investors looking to forecast exchange rates, and banks looking to hedge their risks. Volatility forecasting is challenging because the phenomenon of conditional variance is unobserved and this makes the evaluation and comparison of various proposed models difficult. A great number of models have been proposed since the introduction of the seminal papers by Engle (1982) and Bollerslev (1986) \cite{Engle1982,Bollerslev1986}. DeGooijer provides a comprehensive overview of the various models that have come to the fore over the last 25 year \cite{DeGooijer2006}. Most of the models that are extensively studied and used in practice are parametric in nature and only recently has there been a move towards exploring semi-parametric and non-parametric models. An exploration of various parametric and non-parametric methods of volatility measurement has been summarised in literature \cite{Andersen2005}, as well as the desired properties of good volatility forecasts \cite{Armstrong2001,Poon2005}. 

The most widely used parametric model for volatility forecasting is the Generalized Autoregressive Conditional Heteroskedascity (GARCH) model, introduced by Bollerslev in 1986, which is an offshoot of the ARCH model introduced by Engle in 1982. GARCH assumes a linear relationship between the current and past values of volatility, and assumes that positive and negative returns have the same effect on volatility. Many variations have been introduced to the concept of the GARCH model, most of which try to address one or both of these assumptions, each with its own merits and demerits \cite{Hansen2005}. 

This paper presents volatility forecasting methods based on Bayesian non-parametrics (using Gaussian Processes) and provides a comparison with traditional approaches where possible. The use of Gaussian processes for volatility forecasting has been explored preliminarily in literature \cite{Dorion2013,Platanios2013}. In this paper we extend this initial work to consider the performance of Gaussian Processes in capturing well known aspects of financial volatility; namely, volatility clustering, asymmetric effect of positive and negative returns on volatility, and a robustness to outliers.  

This paper is organized as follows. Section \ref{sec:ret} defines the financial variables of interest, Section \ref{sec:GARCH} describes the GARCH model and some of its variants, Section \ref{sec:GP} provides a brief review of Gaussian processes. In Section \ref{sec:trad} we describe traditional approaches used for volatility forecasting, followed in Section \ref{sec:nov} by our approach. In Section  \ref{sec:method} we explain the methodology behind our experiments, the kernels, the loss functions and the accuracy metrics used. In section \ref{sec:exp} we present our data, setup and experiments. Section \ref{sec:sum} presents a summary of the results and conclusions are made in Section \ref{sec:con}.

\section{\scshape Defining Returns}
\label{sec:ret}
Financial time series are non-stationary in the mean, and the linear trend in the mean can be removed by taking the first difference of the time series, as given by 
\begin{IEEEeqnarray}{rCl}
\label{eq:aret}
r_{t} & = & (p_{t} - p_{t-1}) / p_{t-1},
\end{IEEEeqnarray}
where $p_{t}$ and $p_{t-1}$  are the prices at time $t$ and $t-1$ respectively. These are known as arithmetic returns. An alternative to the arithmetic return is to take the geometric returns, also known as the log returns. We obtain the log returns $r_{t}$ by using:
\begin{IEEEeqnarray}{rCl}
\label{eq:gret}
r_{t} & = & \log (p_{t})- \log (p_{t-1}).
\end{IEEEeqnarray}
In this paper we use geometric returns throughout.

\section{\scshape GARCH and Some Variants}
\label{sec:GARCH}
The most commonly used model for financial data is GARCH. It assumes that the returns arise from a zero mean, and heteroskedastic Gaussian. Defining a white noise (Wiener) process $r_t$, as:
\begin{IEEEeqnarray}{rCl}
	\label{eq:dist}
	r_t & \sim & \mathcal{N}(0,{\sigma}_{t}^{2}) 
\end{IEEEeqnarray}
the GARCH($p,q$) model predicts the underlying variance of a time-series as a linear (auto-regressive) combination of $p$ past observed variances and a moving average of the noise process, through $q$ previous squared returns. This linear combination is uniquely defined by coefficient set ${\alpha}_{0:q}$ and ${\beta}_{1:p}$, as:

\begin{IEEEeqnarray}{rCl}
\label{eq:GARCH}
\sigma_{t}^{2} & = & \mathcal{\alpha}_0 + \sum_{j=1}^q \mathcal{\alpha}_j r_{t-j}^2 + \sum_{i=1}^p \mathcal{\beta}_i \sigma_{t-i}^2
\end{IEEEeqnarray}
\\
GARCH, however, suffers from two major drawbacks: first it does not distinguish between the different effects of negative vs positive returns on volatility, second it assumes a linear relationship between the variables. Several mutations of the GARCH have been proposed to overcome these limitations and two of the most commonly used GARCH variants are EGARCH and GJR-GARCH \cite{Hansen2003,Hansen2006}

The exponential GARCH given in 
\begin{IEEEeqnarray}{rCl}
\label{eq:EGARCH}
\log (\sigma_{t}^{2}) & = & \mathcal{\alpha}_0 + \sum_{j=1}^q \mathcal{\alpha}_j g(r_{t-j}) + \sum_{i=1}^p \mathcal{\beta}_i \log (\sigma_{t-i}^2)
\end{IEEEeqnarray}
adds flexibility to the GARCH model by allowing for the asymmetric effect to be captured using $g(x_t)$ where a negative value of ${\theta}$ in 
\begin{IEEEeqnarray}{rCl}
	\label{eq:EGassym}
	g(x_t) & = & \mathcal{\theta} r_t + \mathcal{\lambda} |r_t|
\end{IEEEeqnarray} 
will cause the negative returns to have a more pronounced effect on the volatility.

GJR-GARCH, defined by: 
\begin{IEEEeqnarray}{rCl}
\label{eq:GJR}
\nonumber
\sigma_{t}^{2} & = & \mathcal{\alpha}_0 + \sum_{j=1}^q \mathcal{\alpha}_j r_{t-j}^2 + \sum_{i=1}^p \mathcal{\beta}_i \sigma_{t-i}^2 + \sum_{k=1}^r \mathcal{\gamma}_k r_{t-k}^2 I_{t-k}
\\
I_{t-k} & = & 
\begin{cases}
    0, & \text{if }  r_{t-k} \geq 0\\
    1,& \text{if }  r_{t-k} <0
\end{cases}
\end{IEEEeqnarray}
tries to capture the different effects of positive and negative returns by using a leverage term which only activates in the case of the return being negative.

In the majority of cases, the coefficients of these GARCH models are estimated by using least squares and maximum-likelihood.

\section{\scshape Review of Gaussian Processes}
\label{sec:GP}
A general familiarity with Gaussian processes is assumed, and thus only the information most relevant for the current work is presented. For a detailed treatment of Gaussian processes refer to \cite{Williams1996} and \cite{Rasmussen2006}. 

Gaussian processes form a class of non-parametric models for classification and non-linear regression. They have become widely popular in the machine learning community where they have been applied to a wide range of problems. Their major strength lies in their flexibility and the ease of their computational implementation. 

A Gaussian process is a collection, possibly infinite, of  random variables, any finite subset of which have a joint Gaussian distribution. For a fuction $\mathbf{y} = f(\mathbf{x})$, drawn from a multi-variate Gaussian distribution, where $\mathbf{y} = \{y_1, y_2, ..., y_n \}$ are the values of the depentent variable evaluated at the set of locations $\mathbf{x} = \{x_1, x_2,..., x_n \}$, we can denote this as
\begin{equation}
p(\mathbf{y}) = \mathcal{N}(\mathbf{y}; \boldsymbol{\mu}(\mathbf{x}),\mathbf{K}(\mathbf{x},\mathbf{x}))
\label{eq:mvnorm}
\end{equation}
 where $\boldsymbol{\mu}$ is a \emph{mean function}, and $\mathbf{K}(\mathbf{x},\mathbf{x})$ is the \emph{covariance matrix}, given as
\begin{equation}
\mathbf{K}(\mathbf{x},\mathbf{x}) = \left (
\begin{array}{cccc}
 k(x_1,x_1) & k(x_1,x_2) & \cdots & k(x_1,x_n)\\
 k(x_2,x_1) & k(x_2,x_2) & \cdots & k(x_2,x_n)\\
 \vdots & \vdots & \vdots & \vdots \\
 k(x_n,x_1) & k(x_n,x_2) & \cdots & k(x_n,x_n)
\end{array}
\right )
\label{eq:covMat}
\end{equation}
Each element of the covariance matrix is given by a function $k(x_i, x_j)$, which is called the \emph{covariance kernel}. This kernel gives us the covariance between any two sample locations, and the selection of this kernel depends on our prior knowledge and assumptions about the observed data.

In order to evaluate the Gaussian process posterior distribution at a new test point $x_*$ we use the joint distribution of the observed data and the new test point,
\begin{equation}
p \left ( \left [
 \begin{array}{c}
 \mathbf{y}\\
  y_*
 \end{array}
\right] \right ) =
\mathcal{N} \left (
\left [
 \begin{array}{c}
  \boldsymbol{\mu}(\mathbf{x}) \\
  \mu(x_*)
 \end{array}
\right ] ,
\left [
\begin{array}{cc}
 \mathbf{K}(\mathbf{x},\mathbf{x}) & \mathbf{K}(\mathbf{x},x_*)\\
 \mathbf{K}(x_*,\mathbf{x}) & k(x_*,x_*)
\end{array}
\right ]
\right )
\label{eq:gppost1}
\end{equation}
where $\mathbf{K}(\mathbf{x},x_*)$ denotes a column vector comprised of $k(x_1,x_*),...,k(x_n,x_*)$ and $\mathbf{K}(x_*, \mathbf{x})$ is its transpose. By matrix maniputlation, we find that the posterior distribution over $y_*$ is Gaussian and its mean and variance are given by
\begin{equation}
m_* = \mu(x_*) + \mathbf{K}(x_*,\mathbf{x}) \mathbf{K}(\mathbf{x},\mathbf{x})^{-1} (\mathbf{y}-\boldsymbol{\mu}(\mathbf{x})) \text{\quad and}
\label{eq:GPm}
\end{equation}
\begin{equation}
\sigma_{*}^{2} = K(x_*,x_*) - \mathbf{K}(x_*,\mathbf{x}) \mathbf{K}(\mathbf{x},\mathbf{x})^{-1} \mathbf{K}(\mathbf{x},x_*).
\label{eq:GPv}
\end{equation}
This can be extended for a set of locations outside our observed data set, say $\mathbf{x}_*$, to find the posterior distribution of $\mathbf{y}_*)$. Using standard results for multivariate Gaussians, the extended equations for the posterior mean and variance are given by
\begin{equation} \label{eq:GPMeanVar}
p(\mathbf{y}_* \mid \mathbf{y}) = \mathcal{N}(\mathbf{y}_*; \mathbf{m}_*, \mathbf{C}_*)
\end{equation}
where,
\begin{eqnarray}
\mathbf{m}_* = \boldsymbol{\mu}(\mathbf{x}_*) + \mathbf{K}(\mathbf{x}_*, \mathbf{x})\mathbf{K}(\mathbf{x},\mathbf{x})^{-1}(\mathbf{y}(\mathbf{x})-\boldsymbol{\mu}(\mathbf{x})) \label{eq:GPMean}\\
\mathbf{C}_*  = \mathbf{K}(\mathbf{x}_*, \mathbf{x}_*)-\mathbf{K}(\mathbf{x}_*,\mathbf{x})\mathbf{K}(\mathbf{x},\mathbf{x})^{-1}\mathbf{K}(\mathbf{x}_*,\mathbf{x})^\top \label{eq:GPVar}\,.
\end{eqnarray}
If the observed function values, $y_i$, have noise associated with them, then we can bring a noise term into the covariance. Since the noise of each sample is expected to be uncorrelated therefore the noise term only adds to the diagonal of $\mathbf{K}$. The covariance for noisy observations is given as
\begin{equation}
\mathbf{V}(\mathbf{x},\mathbf{x}) = \mathbf{K}(\mathbf{x},\mathbf{x}) + \sigma_n^2 \mathbf{I},
\label{eq:addnoise}
\end{equation}
where $\mathbf{I}$ is the identity matrix and $\sigma_n^2$ is a \emph{hyperparameter} for the noise variance.
For finding the posterior distributions of new test points for noisy data, we simply replace the $\mathbf{K}(\mathbf{x},\mathbf{x})$ term in the above equations with, $\mathbf{V}(\mathbf{x},\mathbf{x})$ from Eq. \ref{eq:addnoise}.
\\
\section{\scshape Gaussian Processes for Volatility Estimation}
\label{sec:trad}
\begin {comment} {Extensive literature exists, suggesting and positing the merits of various measures of the latent variance.} 
\end{comment} 
Given a financial time series, if we assume a Gaussian prior for the returns, i.e. the returns are individually drawn from a normal distribution, as given by Eq. \eqref{eq:dist}, then the  variance of the normal distribution at that point is a measure of the volatility. If we fit a Gaussian process to the data, then our best estimate of the value of the return at the next time step is the posterior mean of the Gaussian process, as given by Eq. \eqref{eq:GPm}, and the forecasted variance, given by Eq. \eqref{eq:GPv}, is our best estimate of the volatility for the future time step. Practically, when we work directly with returns, we find that regressing through the data points using a stationary Gaussian process results in a very smooth mean estimation, with large and mostly unvarying variance bounds, giving us very little information about of the underlying volatility function. Although a GP fitted to the returns can provide useful information like helping us to determine mean-reversion cycles, or market bias etc., but we are interested mainly in the variance, or envelope function of the returns. 

In most volatility models, some transformation of the financial returns is used as a proxy for the latent variance. Usually squared or absolute returns are used in much of the time series literature though absolute returns have been shown to be a more robust measure \cite{Poon2003}. In this paper, we use the absolute returns as a proxy for the underlying variance, when carrying out the comparison of GPs and GARCH. The variance can be obtained by predicting the time series of $|r|$ or $r^2$.
\section{Extensions to the Gaussian Process Model}
\label{sec:nov}
Noting that volatility, defined via absolute returns, is a strictly positive quantity, we perform GP regression on the log-transformed absolute returns. This has the major advantage of enforcing the positivity constraint on our solutions, whilst retaining a standard GP in the log-space. Predictive measures of uncertainty in the log-space are then readily transformed back, noting that they form asymmetric bounds. 

\subsubsection{Regressing in Log-space: }
Defining our target variables $y$ as $y = \log(|r_t|)$, or as $y = \log(r^2_t)$ when working with squared returns, we may readily perform GP regression over the set of observed $y$. It remains a simple task then to transform the next step forecast and credibility intervals of the predictive distribution on $y$ to ones on $|r|$ using $\exp()$ to invert our log transform as: 
\begin{IEEEeqnarray}{rCl}
\label{eq:Norm}
\bar r_* & = & \exp(\bar f_{*})
\end{IEEEeqnarray}
and the upper and lower intervals can be recovered as: 
\begin{IEEEeqnarray}{rCl}
\label{eq:upcon}
c_{up} & = & \exp(\bar f_{*} + 1.96\sigma_*)
\\
\label{eq:dncon}
c_{low} & = & \exp(\bar f_{*} - 1.96\sigma_*)
\end{IEEEeqnarray}
Here we choose 95\% intervals, readily obtained from scaling the predictive standard deviation on $y$ by 1.96, and $\sigma_{*}^{2} = \mathbb{V}[f_{*}]$ from Eq. \eqref{eq:GPv}.

\subsubsection{Separating Positive and Negative Returns: }
\begin{figure*}[tb]
	\centering
	\includegraphics[width=\textwidth, trim = 55mm 0mm 40mm 0mm, clip]{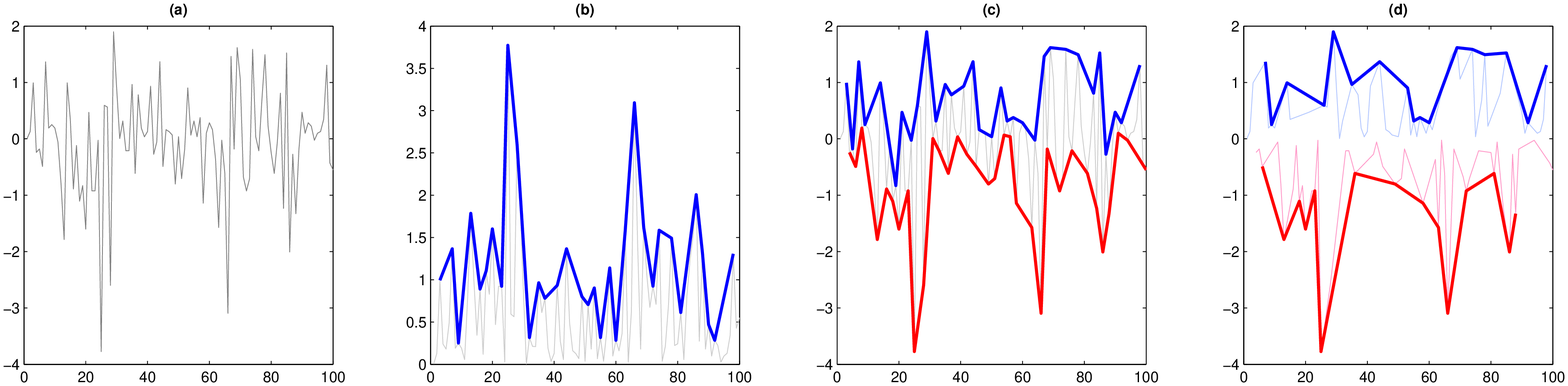}
	\caption{Magnitude of returns on the vertical axis with time on the horizontal (a) Geometric returns. (b) Absolute returns and the maximum envelope function. (c) Positive (blue) and negative (red) maximum envelopes of un-separated Geometric returns. (d) Separated positive (blue) and negative (red) returns and their maximum envelops }
	\label{fig:Approaches}
\end{figure*}

Simply using absolute or squared returns does not address the observation made earlier that negative and positive returns affect the volatility differently. To capture this effect, we treat the positive and negative returns as two data series $g_{+}$ and $g_{-}$ given by 
\begin{IEEEeqnarray}{rCl}
	\label{eq:pnp}
	g_{+}(t) & = &  r_t,\text{ if }  r_{t} \geq 0
	\\
	\label{eq:pnn}
	g_{-}(t) & = & -r_t, \text{ if }  r_{t} <0 
\end{IEEEeqnarray}
and regress on each of these separately in log-space and use the average of the forecasts obtained as our prediction for the next time step, given by:
\begin{IEEEeqnarray}{rCl}
\label{eq:pna}
\bar r_* & = & (\bar r_{+*} + \bar r_{-*}) / 2
\end{IEEEeqnarray}
A sample of separated negative and positive returns can be seen in figure \ref{fig:Approaches}(d).

\subsubsection{Returns Envelope: }
\begin{figure}[tb]
	\scriptsize
	\noindent\begin{minipage}[t]{.24\textwidth}
	\includegraphics[width=\columnwidth, trim = 26mm 149mm 130mm 18mm, clip]{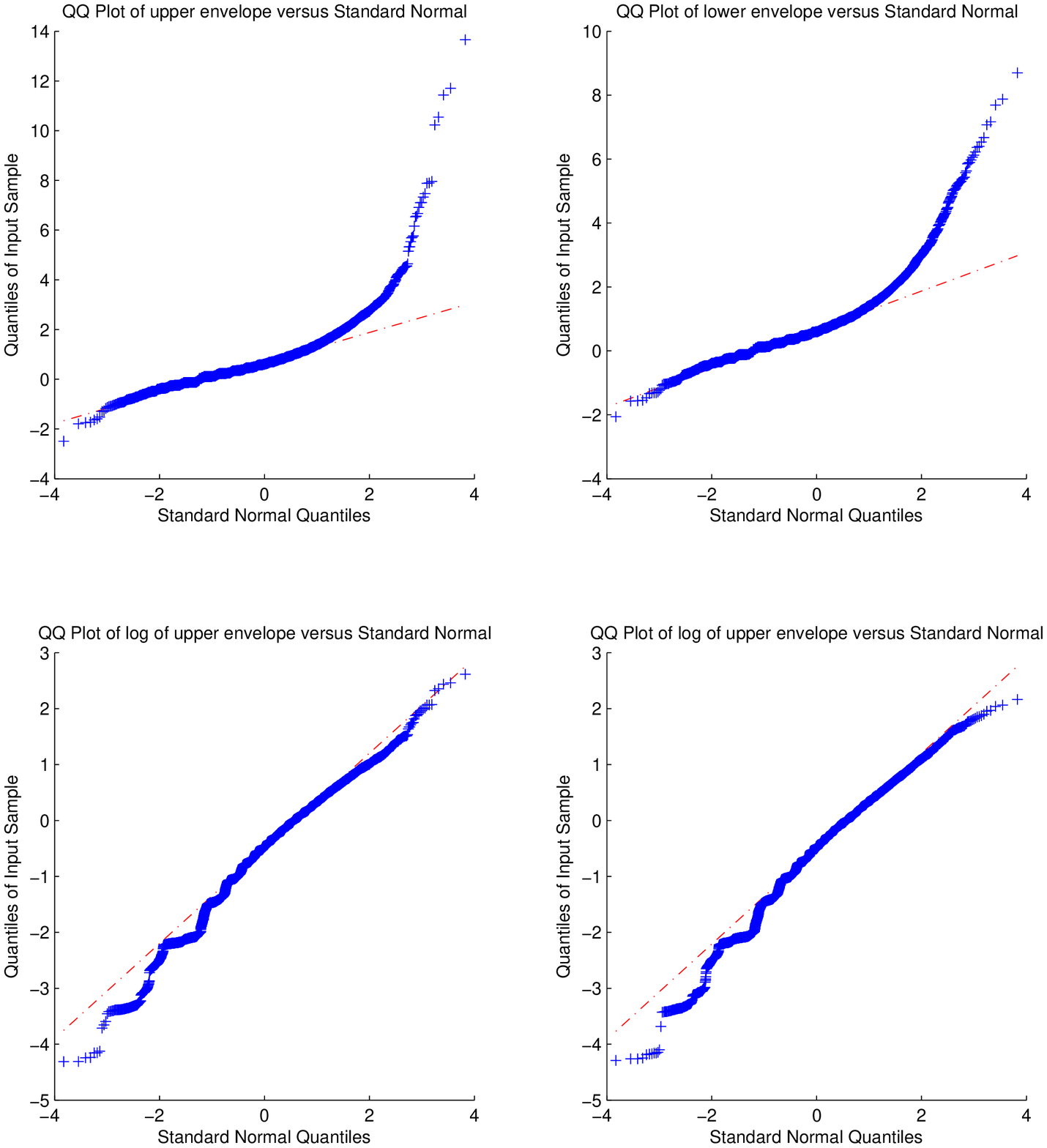}
	\end{minipage}
	\hfill
	\begin{minipage}[t]{.24\textwidth}
	\includegraphics[width=\columnwidth, trim = 136mm 149mm 22mm 18mm, clip]{qqenv.eps}
	\end{minipage}
	\hfill
	\begin{minipage}[b]{.24\textwidth}
	\includegraphics[width=\columnwidth, trim = 26mm 25mm 130mm 142mm, clip]{qqenv.eps}
	\end{minipage}
	\hfill
	\begin{minipage}[b]{.24\textwidth}
	\includegraphics[width=\columnwidth, trim = 136mm 25mm 22mm 142mm, clip]{qqenv.eps}
	\end{minipage}
\caption{QQ plots vs the standard normal, standard normal quantiles on the horizontal axis and quantiles of the input sample on the vertical (a) QQ plot of the maxima (upper) envelope. (b) QQ plot of the minima (lower) envelope. (c) QQ plot of the log of the maxima (upper) envelope. (d) QQ plot of the log of the minima (lower) envelope}
\label{fig:qqenv}
\end{figure}

The envelope of a given time series is a subset of data points that constitutes the maxima or minima of the time series. Whether a point in the data series belongs to the maxima or minima envelope is determined by: \begin{IEEEeqnarray}{rCl}
	\label{eq:envu}
	z_{max} & = & g_t,  \text{ if }  g_{t} \geq g_{t-1} \text{ and } g_{t} \geq g_{t+1}
	\\
	\label{eq:envd}
	z_{min} & = & g_t,  \text{ if }  g_{t} \leq g_{t-1} \text{ and } g_{t} \leq g_{t+1}
\end{IEEEeqnarray}
For financial time series we can regress on the maxima envelope of the absolute returns,\footnote{When we say absolute returns, it can be assumed that the same applies for the squared returns.} shown in figure \ref{fig:Approaches}(b). We can also regress on the maxima envelope of the positive returns, the minima envelope of the negative returns, shown in figure \ref{fig:Approaches}(d), and then combine the results by using Eq. \eqref{eq:pna}. The envelopes of financial time series display a log-normal distribution as can be seen in figure \ref{fig:qqenv}, and GP regression on envelopes maintains the assumption in Eq. \eqref{eq:dist}. When regressing on the envelope, we are dealing with a lesser number of data points (practically about only one-third of the original data series), and our regression is less affected by the intermediate fluctuation of the data points and we get a relatively smoother estimate of the underlying volatility function.

\section{\scshape Methodology}
\label{sec:method}
Our aim is to evaluate the performance of volatility forecasting approaches based on Gaussian processes, against existing approaches such as GARCH and its commonly used variants. In this section we outline the measures we use for evaluating the forecasting performance. We explain our approach for selecting the best GP kernel for our data, outline the technique we use for hyperparameter inference and our use of Cholesky updating and downdating for updating our covariance matrix at each forecasting step. We also establish a method to ensure that GP forecasts of volatility are an unbiased measure of the underlying variance by comparing the residuals obtained by GARCH versus those obtained from a GP. 

\subsubsection{Performance Metrics: }
The forecasting performance of the prediction technique is evaluated by comparing the underlying volatility $\sigma^{2}$, in this case the proxy for volatility i.e.\ absolute or squared returns, to the one step ahead forecasts $h$. For evaluating the performance of the different kernels we use four loss functions and two accuracy measures. The loss functions are the same as used by \cite{Hansen2005}, and are extensively used in financial volatility forecasting literature for comparison of predicting ability of different models:
\begin{IEEEeqnarray}{rCl}
MSE_{1} & = & n^{-1}\sum_{t=1}^n(\sigma_{t}-h_{t})^{2}
\\
MSE_{2} & = & n^{-1}\sum_{t=1}^n(\sigma_{t}^{2}-h_{t}^{2})^{2}
\\
MAE_{1} & = &  n^{-1}\sum_{t=1}^n|\sigma_{t}-h_{t}|
\\
MAE_{2}  & = &  n^{-1}\sum_{t=1}^n|\sigma_{t}^{2}-h_{t}^{2}|
\\
\end{IEEEeqnarray}
The accuracy measures are two of the five that are used in the m-competition \cite{Makridakis2000}, namely \textit{Median RAE (relative absolute error)}, and \textit{Symmetric mean absolute percentage error (sMAPE)}: 
\begin{IEEEeqnarray}{rCl}
MdRAE  & = & \text{median}\left(\left|\frac{\sigma_{t}-h_{t}}{\sigma_{t}-\sigma_{t-1}}\right|\right)
\\
sMAPE  & = &  n^{-1}\sum_{t=1}^n\left(\frac{|\sigma-h|}{\sigma+h}\right)*200
\end{IEEEeqnarray}
The MdRAE measures the error for the proposed model against the no-change model, in which our one step ahead forecast is the same as the last observed data point. A value of less than 1 means that the proposed model performs better than the naive no-change model. For more information on this measure, see \cite{Armstrong1992}.

The sMAPE, gives symmetric weightage to over and under predicted values if they are the same absolute distance from the original value, see \cite{Makridakis2000} for details.

\subsubsection{Kernel Selection: }
When applying GPs to financial forecasting, a primary consideration is that selection of an appropriate kernel to best suit our data. This establishes our prior knowledge about the function space. In this paper we compare three different kernels: the squared exponential kernel, the Mat\'{e}rn-3/2 kernel, and the quasi-periodic kernel outlined in \cite{Roberts2013}. In our kernel comparison, the SE kernel serves as the baseline against which we compare the performance of the other two kernels.

Let $d = |x_{i}-x_{j}|$, then:
\begin{IEEEeqnarray}{rCl}
k_{SE} & = & \sigma_{h}^{2} \exp\left[-\left(\frac{d}{\sqrt{2} l}\right)^{2}\right]
\\
k_{\text{Mat\'{e}rn}, \nu=3/2} & = & \left(1+\frac{\sqrt{3}d}{l}\right) \exp\left[-\left(\frac{\sqrt{3}d}{l}\right)\right]
\end{IEEEeqnarray}
\\
where $\sigma_{h}$ is the output scale, $l$ is the input length scale and $\sigma_{n}^2$ is the noise variance:
\begin{IEEEeqnarray}{rCl}
k_{QP} & = & \sigma_{h}^{2}\exp\left(-\frac{\sin^{2}[\pi d/T]}{2 w^2}-\frac{d^{2}}{l^{2}}\right)+I\sigma_{n}^2
\end{IEEEeqnarray}
\\
where $T$ is the period and $w$ is a hyperparameter meant to capture the roughness relative to the period, for a further discussion of this kernel, see \cite{Roberts2013}.

\subsubsection{Regression: }
In this paper we use the absolute returns, the envelope of the absolute returns, the separate negative and positive returns as well as their envelopes. We perform GP regression over the data in log-space, using a rolling window approach with data dropping from the tail, as new data comes in. We using the forecasting equations outlined in section 3, and section 4.  

\subsubsection{Hyperparameters Inference: }
We infer the hyperparameters by using a maximum a posteriori probability (MAP) estimate, placing a Gaussian prior distribution over our hyperparameter space. We implement the search over the hyperparamenter space by minimizing the value of the negative log posterior, using the Nelder-Mead simplex (direct search) method \cite{fmin2013}. 

In order to remove the effect of the initial guess affecting the final hyperparameters discovered, a uniformly random grid of $1000^{3}$ initializing points is set up. The grid points serve as the starting guess points from which we explore the hyperparameter space. The hyperparameters which produce the lowest value for the negative log posterior are kept as the best estimate. 

\subsubsection{Cholesky Factor Up-dating and Down-dating: }
Since we are dealing with long financial time series, instead of recomputing our covariance matrix at each forecasting time step, we use Cholesky updating and Cholesky downdating \cite{Osborne2008} to update our covariance matrix. This ensures that after the initial learning phase has been carried out, all future updates occur at a rapid pace making the technique extremely suitable for online use, with live feed data.  

\subsubsection{Hyperparameter Updating: }
We can choose to re-evaluate the hyperparameters at each forecasting step, using our new window of points to conduct the inference. To update our hyperparameters in a rolling window approach, we use the hyperparameters from the previous time step as our best guess for inferring the next set of hyperparameters. This makes the exploration of the parameter space rapid and less prone to high jumps in values of parameters, unless mandated by a change in the nature of the underlying data. 

\section{Experimental Findings}
\label{sec:exp}
In the first set of experiments, we compare of the three kernels mentioned in section IV, against each other to find the best performing kernel. The best performing kernel is then used in the second set of experiments, for comparing the predicting performance of GPs against GARCH. The comparison of GPs vs GARCH is carried out for absolute returns, squared returns, envelope of absolute returns, and the negative and positive returns being treated separately. 
\subsubsection{Data sets: }
Four currency pairs are examined in the experiments. They are the U.S. dollar/Japanese yen pair (USD/JPY), the British pound/U.S. dollar pair (GBP/USD), the Euro/Swiss franc pair (EUR/CHF) and the Euro/U.S. dollar pair (EUR/USD). The whole set comprises of the period from 15 Sept, 2003 to 15 Sept, 2005, with data points at 30 minute intervals. A subset of all available data is used in each run of the experiment, to limit computing overhead, and to find the average performance across different periods. 
\subsubsection{Experimental Protocol: }
Each currency pair data set of 25120 data points is split into groups of 3140 points, corresponding to a period of 13 weeks i.e. a quarter of a year. The data points from the first two days of each group, 100 points, are then used for training and determining the hyperparameters, while the rest of the points are used for one step ahead, out-of-sample forecasting. The regression is carried out in log-space and at each point the mean of the GP regression is taken as the one-step ahead forecast i.e.\ $h_{t+1}$. The actual value of the data point is then observed and made part of the known data set, and then at time $t+1$, the next forecast, $h_{t+2}$, is made. The forecasts are transformed back from log-space and compared to the original data points and forecasting accuracy is measured using the metrics described earlier. To ensure that the inference algorithm is able to consistently uncover the underlying hyperparameters, we test our algorithm on a series of randomly generated data sets, with known hyperparameters.

We generate a random set of hyperparameters (the output length scale, the input scale and the noise variance) and from these we generate 100 data sets with 1000 points each, this is treated as the underlying (actual) function. From the generated data a subset is selected to be used as our observations of the actual process. The subset is selected by picking up a random number of uniformly distributed points from the complete dataset, anywhere from 5\% to 95\% of the data set is used. The hyperparemeters discovered in the previous step are used for carrying out regression on the observed data points. The generated function is then compared with the underlying function. It is found that the hyperparameters can be discovered consistently by using as few as 20\% of the actual data points as observed data. We carry out a set of experiments using the Gaussian processes to determine if using a larger training set affects the performance but a consistent performance improvement is not found as we increase the size of the training data set.
\subsubsection{Kernel Performance: }
The kernels are used to carry out regression on all four data sets, and the Mat\'{e}rn 3/2 kernel is found to give the best results on our performance criteria. The performance of the three kernels, as measured by the eight metrics, for the USD/JPY pair is shown in Fig. \ref{fig:UJ_lossfunctions}. We conduct all further experiments using the Mat\'{e}rn 3/2 kernel.
\subsubsection{Bias in Gaussian Processes: }
\begin{figure*}
	\scriptsize
	\noindent\begin{minipage}[t]{.6\textwidth}
		\nonumber
	\includegraphics[height=4cm, trim = 38mm 14mm 8mm 12mm, clip]{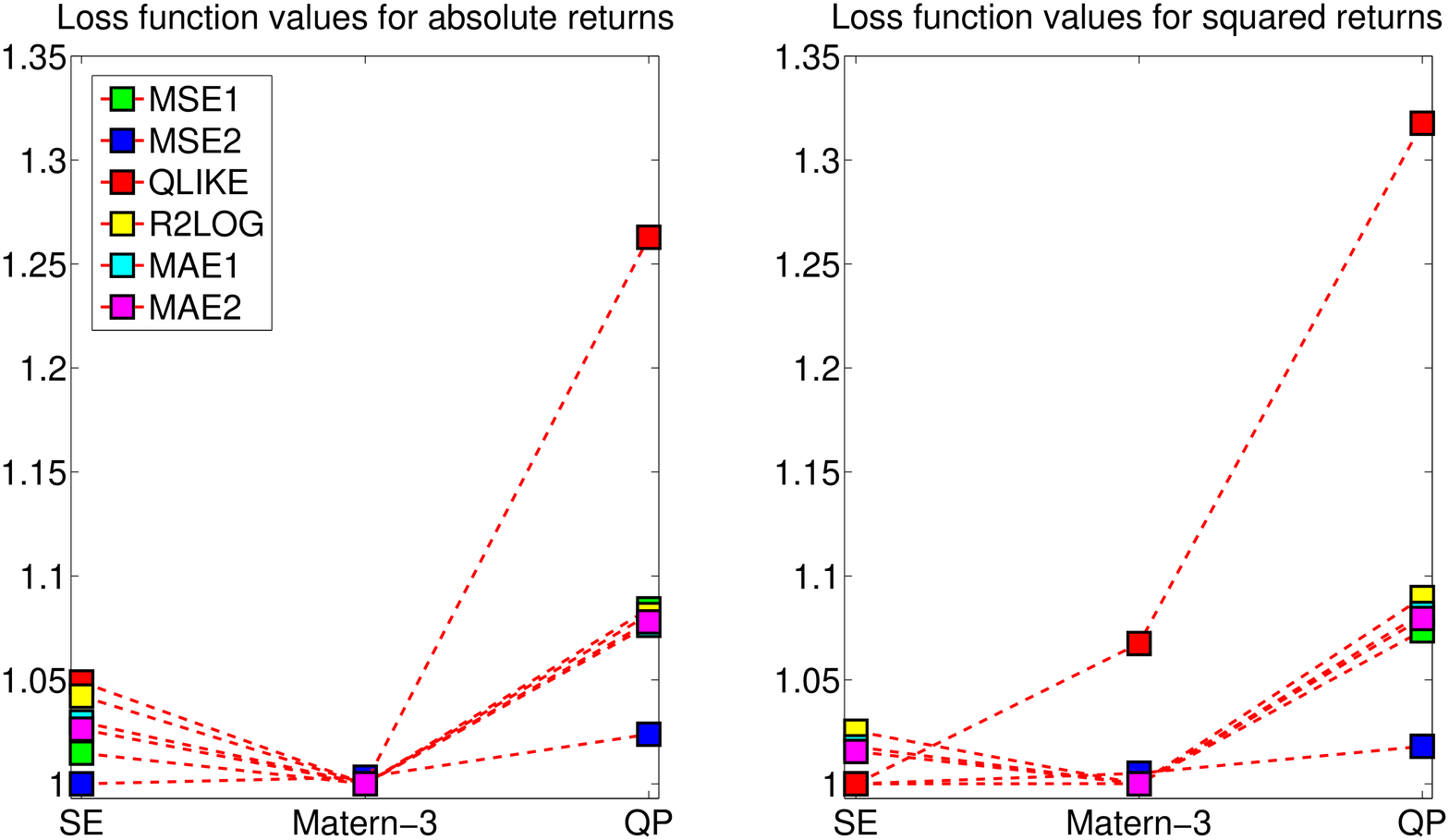}
	\end{minipage}
	\hfill
	\begin{minipage}[t]{\textwidth}
		\nonumber
		\includegraphics[height=3.95cm, trim = 10mm 9mm 8mm 8mm, clip]{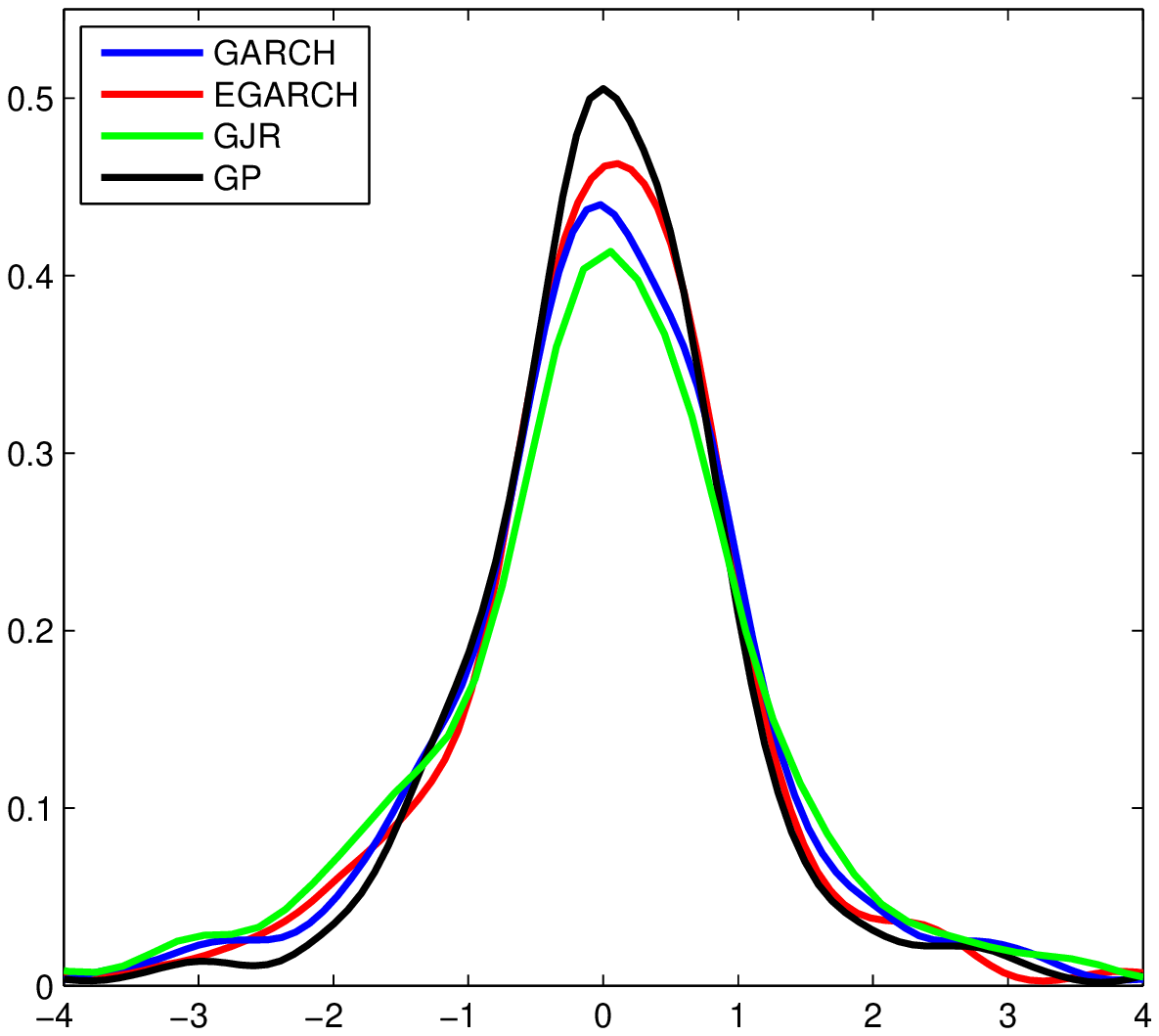}
		\end{minipage}

\begin{minipage}[t]{\textwidth}
	\caption{Left \& centre: Loss function performance of the three kernels for USD/JPY data. Left: for absolute returns, Centre: for squared returns.}
	\label{fig:UJ_lossfunctions}
\end{minipage}
\hfill
	\begin{minipage}[t]{\textwidth}
	\caption{Right: Probability distribution of the residuals that are obtained from normalizing the returns by the forecasts produced by GP regression. The residuals produced by GARCH and its variants are provided for comparison to establish unbiasedness of GPs for volatility prediction}
	\label{fig:Res}
\end{minipage}
\end{figure*}

To establish that forecasts produced by a given method are unbiassed, we carry out element wise normalization of the actual returns $r_t$, using the forecasted time series $f_t$, as given by:
\begin{IEEEeqnarray}{rCl}
	\label{eq:resF}
	e_t & = & r_t / f_{t}
\end{IEEEeqnarray}
The residuals $e_t$ obtained from the normalization should be close to a white noise process i.e.\ it should be Gaussian distributed with zero mean and unit standard deviation given by:
\begin{IEEEeqnarray}{rCl}
	\label{eq:resN}
	e_t & \sim & \mathcal{N}(0,1) 
\end{IEEEeqnarray}
GARCH and its variants are well established to produce Gaussian residuals. The distribution of residuals obtained from GP regression is shown in figure \ref{fig:Res}. The distributions of the residuals of GARCH, EGARCH and GJR-GARCH are also shown for comparison. 

\section{Comparison with GARCH}
\label{sec:sum}
To compare the performance of the Gaussian process against a standard baseline, we use GARCH, EGARCH and GJR-GARCH implementations found in the Matlab Econometrics Toolbox.

As outlined in the general setup of the experiments, 100 data points are used as the training set for determining the parameters of GARCH and its variants. This same training set is used for inferring the hyperparameters of the GPs. At each time step, a one step ahead forecast is carried out using GP approaches, and GARCH. The actual value at the time step is then observed and made part of the data set. This new data set is then made available to both GP and GARCH to make the next time step prediction. The residuals of the GARCH and the GP models are compared to ensure unbiasedness. The predications are compared with the observed values at each time step using the performance metrics. In this way we carry out GP regression on the absolute and squared returns and compare it to the GARCH baseline. We carry out the same set of experiments using the envelopes of the absolute and square returns, with and without hyperparameter updating.

\subsection{\scshape Summary of Results}
The following main results are found from our experiments. 
These results are discussed in some more detail in the following sections.:
\begin{enumerate}
	\item Gaussian process based forecasting approaches consistently perform better than GARCH and its variants.
	\item Gaussian processes give unbiased predictions of volatility as evidenced by the distribution of their residuals.
	\item Envelope based approaches perform better than absolute returns based approaches.
	\item Hyperparameter updating at each time step, gives better prediction performance than retaining one time evaluation of hyperparameters.
	\item Gaussian processes give asymmetric variance bounds for prediction which are better able to explain the variance of the underlying time series, than GARCH.
\end{enumerate}

\subsubsection{Kernel Performance: }
The kernel comparison shows that the Mat\'{e}rn-3/2 gives the best perfomance. The results for the performance of the three different kernels are collated in the four smaller tables, each corresponding to a separate currency pair. Table \ref{UJ_table} shows the forecasting performance of the three kernels on USD/JPY pair, for both absolute and squared returns, as measured by the 6 performance criteria. It can be seen that generally the Mat\'{e}rn-3/2 gives the best performance out of the three kernels, while the quasi-periodic kernel performs the worst, as can be seen in figure \ref{fig:UJ_lossfunctions}. Also it can be seen that the performance of the kernels is not significantly or consistently better in the case of absolute returns versus squared returns as a proxy for volatility. The sMAPE values show that overall the kernels perform 22.5\% better, on average for USD/JPY, than a random walk model. Table \ref{GU_table}, \ref{EU_table}, \ref{EC_table} show the values of the performance measures for GBP/USD, EUR/USD and EUR/CHF respectively. 

The overall sMAPE results demonstrate that for the data under consideration, across all kernels, and for both absolute and squared returns, and envelopes, the models on average perform 21\% better than a random walk model approach. 

\begin{table*}[t]
	\scriptsize
	\noindent\begin{minipage}[t]{0.48\textwidth}
		\caption{Forecasting performance on USD/JPY}
		\begin{tabular}{l c c c|c c c}
			\toprule
			\multicolumn{1}{l}{ } & \multicolumn{3}{c|}{Absolute Returns} & \multicolumn{3}{c}{Squared Returns}\\
			& SE & M-3 & QP & SE & M-3 & QP\\
			\midrule
			MSE1 & 0.57 & \textbf{0.56} & 0.61 & 0.56 & {0.56} & 0.6\\
			MSE2 & \textbf{10.20} & 10.23 & 10.44 & \textbf{10.23} & 10.28 & 10.41\\
			MAE1 & 0.48 & \textbf{0.47} & 0.51 & 0.47 & \textbf{0.46} & 0.5\\
			MAE2 & 1.01 & \textbf{0.98} & 1.06 & 0.99 & \textbf{0.97} & 1.05\\
			MdRAE & 0.82 & \textbf{0.80} & 0.86 & 0.82 & \textbf{0.81} & 0.86\\
			sMAPE & 77.58 & \textbf{76.36} & 79.5 & 76.47 & \textbf{75.93} & 79.39\\
			\bottomrule
		\end{tabular}
		\label{UJ_table}
	\end{minipage}
	\hfill
	\begin{minipage}[t]{0.48\textwidth}
		\caption{Forecasting performance on GBP/USD}
		\begin{tabular}{l c c c|c c c}
			\toprule
			\multicolumn{1}{l}{ } & \multicolumn{3}{c|}{Absolute Returns} & \multicolumn{3}{c}{Squared Returns}\\
			& SE & M-3 & QP & SE & M-3 & QP\\
			\midrule
			MSE1 & 0.52 & 0.52 & 0.52 & 0.52 & 0.52 & 0.52\\
			MSE2 & 8.80 & 8.85 & \textbf{8.79} & 8.83 & 8.91 & 8.83\\
			MAE1 & 0.5 & \textbf{0.48} & 0.5 & 0.49 & \textbf{0.48} & 0.5\\
			MAE2 & 0.98 & \textbf{0.96} & 0.99 & 0.97 & \textbf{0.95} & 0.98\\
			MdRAE & 0.86 & \textbf{0.84} & 0.86 & 0.84 & \textbf{0.82} & 0.85\\
			sMAPE & 77.11 & \textbf{76.06} & 77.68 & 76.53 & \textbf{75.51} & 77.23\\
			\bottomrule
		\end{tabular}
		\label{GU_table}
	\end{minipage}
	\\
	
	\begin{minipage}[t]{0.48\textwidth}
		\caption{Forecasting performance on EUR/USD}
		\begin{tabular}{l c c c|c c c}
			\toprule
			\multicolumn{1}{l}{ } & \multicolumn{3}{c|}{Absolute Returns} & \multicolumn{3}{c}{Squared Returns}\\
			& SE & M-3 & QP & SE & M-3 & QP\\
			\midrule
			MSE1 & 0.57 & \textbf{0.56} & 0.58 & 0.57 & \textbf{0.55} & 0.58\\
			MSE2 & 12.42 & \textbf{12.41} & 12.42 & \textbf{12.41} & 12.42 & 12.42\\
			MAE1 & 0.55 & \textbf{0.53} & 0.56 & 0.55 & \textbf{0.53} & 0.56\\
			MAE2 & 1.07 & \textbf{1.04} & 1.09 & 1.07 & \textbf{1.03} & 1.09\\
			MdRAE & 0.98 & \textbf{0.93} & 0.99 & 0.98 & \textbf{0.93} & 1\\
			sMAPE & 80.78 & \textbf{79.5} & 81.26 & 80.76 & \textbf{79} & 81.33\\
			\bottomrule
		\end{tabular}
		\label{EU_table}
	\end{minipage}
	\hfill
	\begin{minipage}[t]{0.48\textwidth}
		\caption{Forecasting performance on EUR/CHF}
		\begin{tabular}{l c c c|c c c}
			\toprule
			\multicolumn{1}{l}{ } & \multicolumn{3}{c|}{Absolute Returns} & \multicolumn{3}{c}{Squared Returns}\\
			& SE & M-3 & QP & SE & M-3 & QP\\
			\midrule
			MSE1 & 0.5 & 0.5 & 1.03 & \textbf{0.5} & 0.51 & 2.36\\
			MSE2 & \textbf{5.4} & 5.44 & 14.77 & \textbf{5.44} & 5.48 & 12.79\\
			MAE1 & 0.49 & \textbf{0.48} & 0.71 & 0.49 & \textbf{0.48} & 0.83\\
			MAE2 & 0.94 & \textbf{0.92} & 1.64 & 0.93 & \textbf{0.92} & 3.09\\
			MdRAE & 0.82 & \textbf{0.81} & 1.07 & 0.81 & 0.81 & 1.12\\
			sMAPE & 75.25 & \textbf{75.13} & 90.35 & \textbf{75.33} & 75.41 & 94.68\\
			\bottomrule
		\end{tabular}
		\label{EC_table}
	\end{minipage}
\end{table*}

\subsubsection{Performance of GPs vs GARCH: }
\begin{figure}[t]
	\scriptsize
	\noindent\begin{minipage}[t]{.48\textwidth}
		\includegraphics[width=\textwidth, trim = 25mm 5.5mm 20mm 0mm, clip]{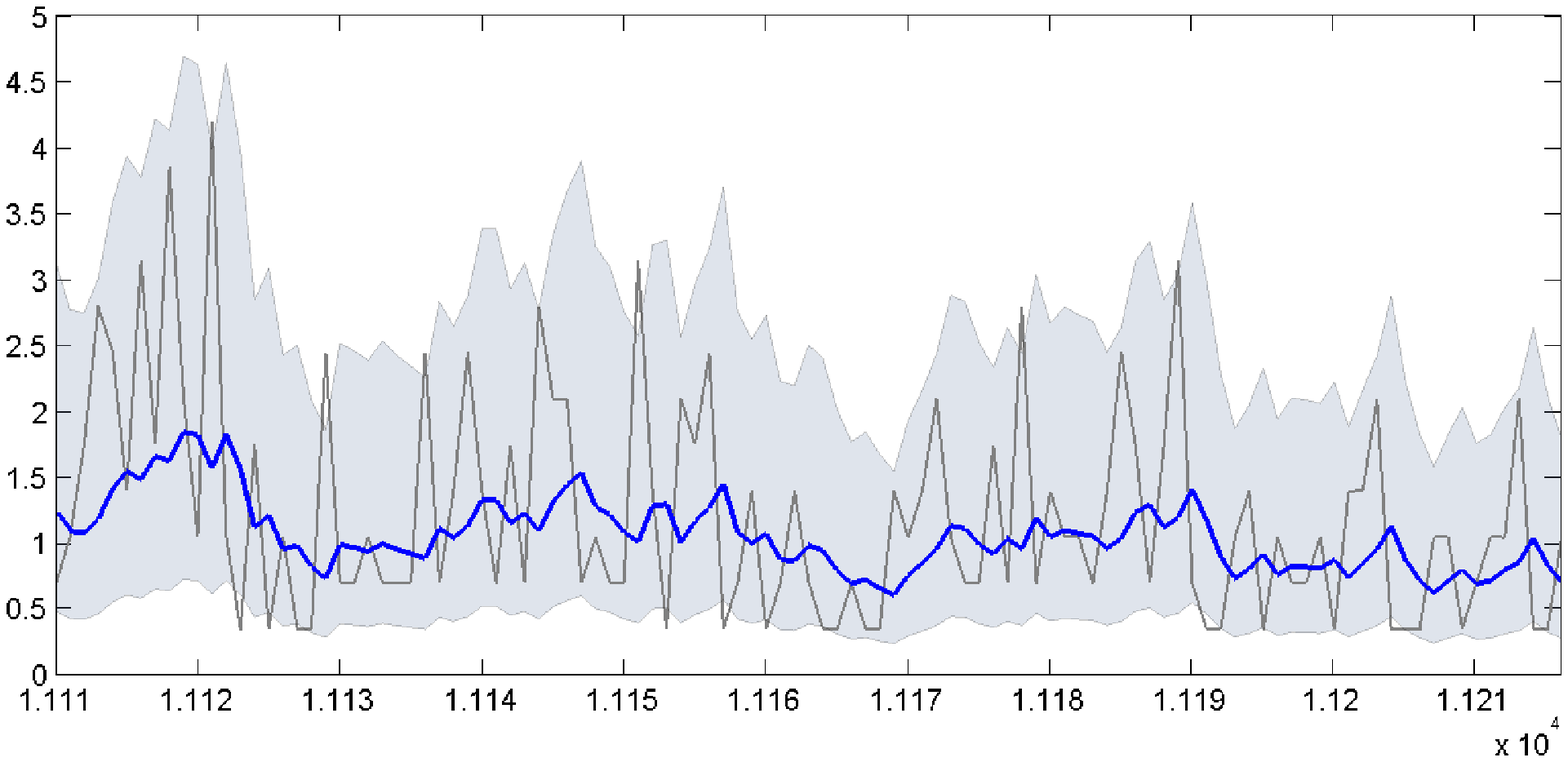}
		\caption{Assymetric variance bounds as observed in normal space, magnitude vs time, obtained from regressing in log-space.}
		\label{fig:varbo}
	\end{minipage}
	\hfill
	\begin{minipage}[t]{.48\textwidth}
		\includegraphics[width=\textwidth, trim = 25mm 5.5mm 20mm 0mm, clip]{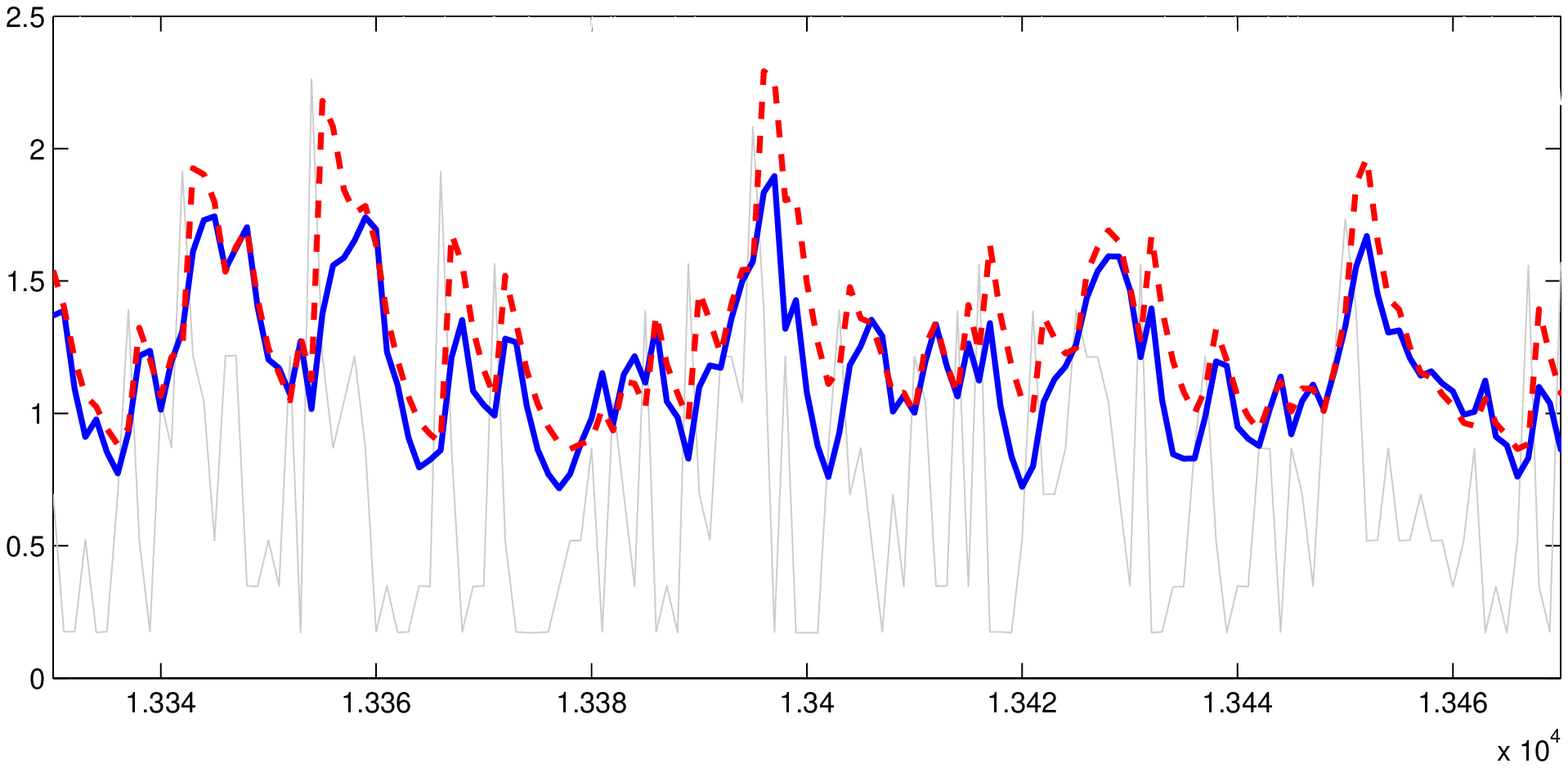}
		\caption{GP Volatility forecasts (blue) obtained by regressing over absolute returns (grey), magnitude vs time, GARCH forecasts (red) are shown for comparison.}
		\label{fig:GPvsGA}
	\end{minipage}
	
	\begin{minipage}[t]{.48\textwidth}
		\includegraphics[width=\textwidth, trim = 25mm 5mm 20mm 0mm, clip]{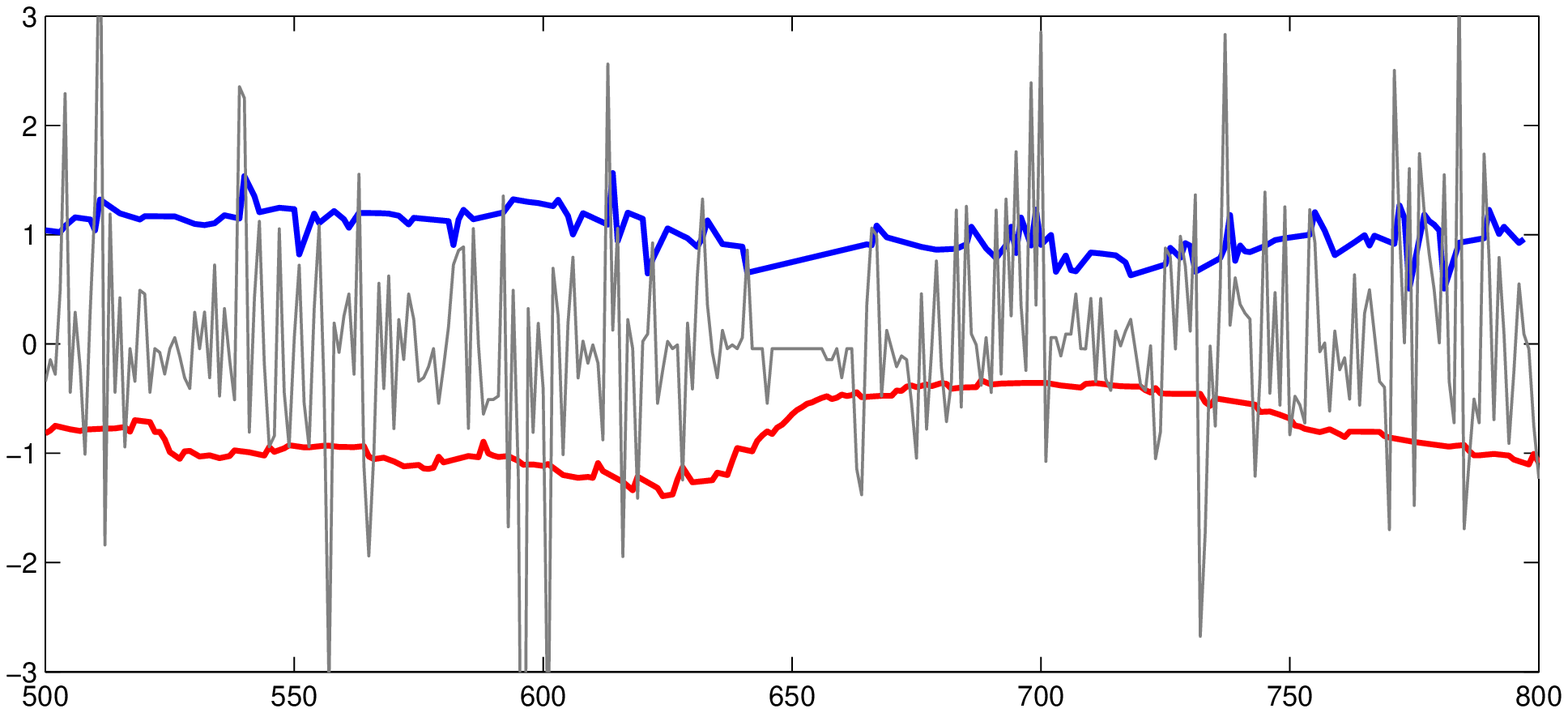}
		\caption{Forecasted mean for Upper \& lower envelopes, magnitude vs time, obtainded by regressing separately on the positive and negative returns.}
		\label{fig:updnenv}
	\end{minipage}
	\hfill
	\begin{minipage}[t]{.48\textwidth}
		\includegraphics[width=\textwidth, trim = 25mm 5mm 20mm 0mm, clip]{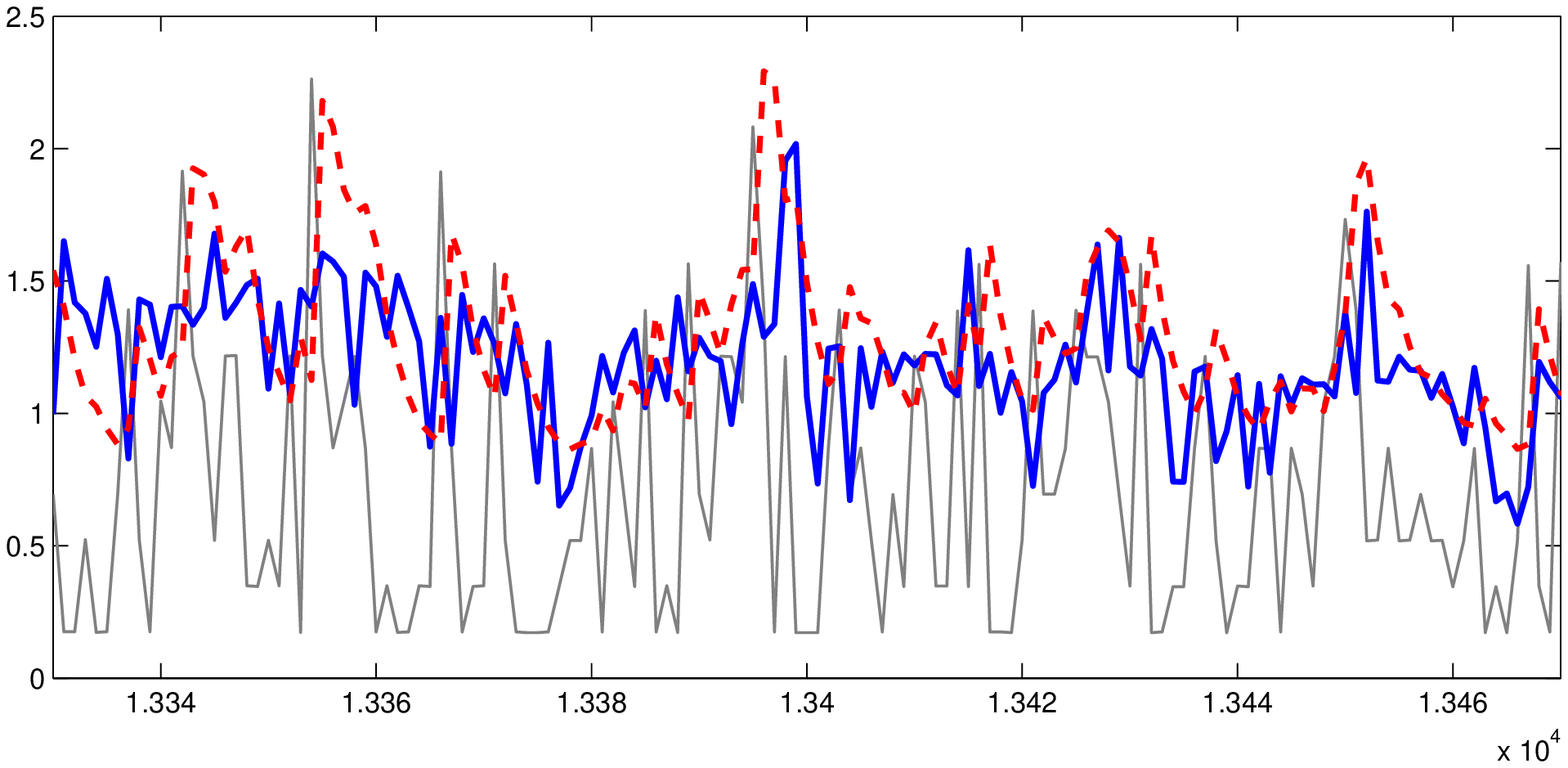}
		\caption{GP Volatility forecasts obtained by combining the regressions from the maxima envelope of the positive returns and the minima envelope of the negative returns, magnitude vs time, GARCH forecasts (red), and absolute returns (grey) are shown for comparison.}
		\label{fig:GPcvsGA}
	\end{minipage}
\end{figure}

Table \ref{full_table} summarises the volatility forecasting performance of the various approaches we have outlined before to our selected baseline i.e\ GARCH, EGARCH, and GJR-GARCH. The performance of the GPs for the EUR/CHF data series is shown, to give a representative idea of the general forecasting ability of the different approaches. Similar performance is obtained for the other three currency pairs. Figure \ref{fig:varbo} shows the 95\% variance bounds in normal space, that we obtain when we regress using GPs in log-space, as can be seen that the variance bounds are asymetric and capture the envelope and thus the variance of the absolute returns. Figure\ref{fig:GPvsGA} shows a comparison of GARCH and GP predictions. In Table \ref{full_table}, it can be seen that on average, GPs outperform GARCH by 50\% on MSE, and envelope based GP approaches outperform other GP approaches by 20\%, and on 75\% of the shown criteria. 
The approach of using hyperparameter updating, gives a 5\% better forecast than the next best forecast of one step ahead volatility as can be seen in Table \ref{full_tablePU}. 
Figure \ref{fig:updnenv} shows the GP regression performed separeately on the positive and negative retuns and figure \ref{fig:GPcvsGA} shows a comparison of GARCH and prediction obtained from combining the two envelopes.  

\begin{table*}
	\scriptsize
	\begin{minipage}[t]{0.48\textwidth}
		\caption[]{Forecasting performance on EUR/CHF without parameter updating}
		\begin{tabular}{l c c c c}
			\toprule
			& MSE1 & MSE2 & MAE1 & MAE2\\
			\midrule
			GARCH & 1.0937 & 6.2433 & 0.7561 & 0.9929\\
			EGARCH & 0.6756 & 12.7247 & 0.6175 & 1.7952\\
			GJR & 0.6270 & 7.3639 & 0.6173 & 1.2519\\
			\midrule
			Squared returns & 1.0743 & 6.0431 & 0.7459 & 0.9711\\
			Absolute\\ returns & 1.0490 & 6.0056 & 0.7450 & \textbf{0.9685}\\
			Absolute\\ envelope & 0.7195 & 13.5619 & 0.5917 & 1.7312\\
			Combined\\ envelope & \textbf{0.5403} & \textbf{5.3321} & \textbf{0.5544} & 1.062\\
			\bottomrule
		\end{tabular}
		\label{full_table}
	\end{minipage}
	\hfill
	\begin{minipage}[t]{0.48\textwidth}
		\caption[]{Forecasting performance on EUR/CHF with parameter updating}
		\begin{tabular}{l c c c c}
			\toprule
			& MSE1 & MSE2 & MAE1 & MAE2\\
			\midrule
			GARCH & 1.0806 & 6.1859 & 0.7447 & 0.9841\\
			EGARCH & 0.6135 & 12.6121 & 0.6087 & 1.7638\\
			GJR & 0.6171 & 7.2582 & 0.6148 & 1.2396\\
			\midrule
			Squared returns & 1.0190 & 5.6786 & 0.7147 & 0.9179\\
			Absolute\\ returns & 0.9972 & 5.7519 & 0.7121 & 1.0142\\
			Absolute\\envelope & 0.6751 & 12.7272 & 0.5667 & 1.6245\\
			Combined\\envelope & \textbf{0.5113} & \textbf{5.0875} & \textbf{0.5249} & \textbf{0.9205}\\
			\bottomrule
		\end{tabular}
		\label{full_tablePU}
	\end{minipage}
\end{table*}

\section{\scshape Conclusion}
\label{sec:con}
In this paper we have extended Gaussian Process (GP) regression for predicting out-of-sample volatility of financial returns and provided a direct comparison of their performance to traditional approaches such as GARCH, EGARCH, and GJR-GARCH. We compare the forecasting power of three approaches: GP regression on the absolute and squared returns, regression on the envelope of the returns and the absolute returns, and regression on the envelope of the negative and positive returns separately. We use a MAP estimate with a Gaussian prior to determine our hyperparameters. We also test the effect of hyperparameter updating at each forecasting step. We use our approaches to forecast out-of-sample volatility of four currency pairs over a 2 year period, at half-hourly intervals. From three kernels, we select the kernel giving the best performance for our data, which is shown to be Mat\'{e}rn-3/2.  No significant or consistent evidence was found for the superiority of using absolute returns over squared returns, to make gains in terms of predictive ability, as per the metrics used. We use two accuracy measures from the m-competition and 4 statistical loss functions to evaluate the forecasting ability GARCH vs GPs. In general the GPs outperform GARCH by 45\% on MSE, the envelope based GP approaches outperform other GP approaches by 30\%, and the approach of using hyperparameter updating, gives a 5\% improvement in forecasting accuracy.

\bibliographystyle{splncs03.bst}
\bibliography{library}{}

\end{document}